\documentclass[letterpaper, 10 pt, conference]{ieeeconf}  

\IEEEoverridecommandlockouts                              

\usepackage{graphics} %

\usepackage{epsfig} %
\usepackage{mathptmx} %
\usepackage{times} %
\usepackage{amsmath} %
\usepackage{amssymb}  %
\usepackage{multirow}
\usepackage{color}
\usepackage{amsfonts}
\usepackage{xcolor}
\usepackage{tikz}
\usepackage{etex}
\usepackage{url}
\usepackage{footnote}
\usepackage{floatrow}
\usepackage{booktabs}
\usepackage{inputenc}
\usepackage{txfonts}
\usepackage{multicol}
\usepackage{siunitx}
\usepackage{rotating}
\usepackage{arydshln}
\usepackage{algorithm}
\usepackage{algorithmic}
\usepackage{subcaption}
\usepackage[noadjust]{cite} 

\title{\LARGE \bf
A Talent-infused Policy-gradient Approach to Efficient Co-Design of Morphology and Task Allocation Behavior of Multi-Robot Systems
}

\author{Prajit KrisshnaKumar$^{1}$, Steve Paul$^{1}$ and Souma Chowdhury$^{1,2, \dagger}$
\thanks{$^\dagger$ Corresponding Author, soumacho@buffalo.edu}
\thanks{$^{1}$ PhD Student, Mechanical and Aerospace Engineering,
        University at Buffalo, Buffalo, NY }%
\thanks{$^{2}$Associate Professor, Mechanical and Aerospace Engineering,
Co-Director, Center for Embodied Autonomy and Robotics,
University at Buffalo }%
\thanks{*Copyright \textcopyright 2024 Springer. Personal use of this material is permitted. Permission from Springer must be obtained for all other uses, in any current or future media, including reprinting/republishing this material for advertising or promotional purposes, creating new collective works, for resale or redistribution to servers or lists, or reuse of any copyrighted component of this work in other works}
\thanks{*This work was supported by the NSF award CMMI 2048020 and the ONR grant N00014-21-1-2530. Any opinions, findings, conclusions, or recommendations expressed in this paper are those of the authors and do not necessarily reflect the views of NSF and/or ONR.}
}

\begin{document}

\maketitle
\thispagestyle{empty}
\pagestyle{empty}

\begin{abstract}
Interesting and efficient collective behavior observed in multi-robot or swarm systems emerges from the individual behavior of the robots. The functional space of individual robot behaviors is in turn shaped or constrained by the robot's morphology or physical design. Thus the full potential of multi-robot systems can be realized by concurrently optimizing the morphology and behavior of individual robots, informed by the environment's feedback about their collective performance, as opposed to treating morphology and behavior choices disparately or in sequence (the classical approach). This paper presents an efficient concurrent design or co-design method to explore this potential and understand how morphology choices impact collective behavior, particularly in an MRTA problem focused on a flood response scenario, where the individual behavior is designed via graph reinforcement learning. Computational efficiency in this case is attributed to a new way of near exact decomposition of the co-design problem into a series of simpler optimization and learning problems. This is achieved through i) the identification and use of the Pareto front of Talent metrics that represent morphology-dependent robot capabilities, and ii) learning the selection of Talent best trade-offs and individual robot policy that jointly maximizes the MRTA performance. Applied to a multi-unmanned aerial vehicle flood response use case, the co-design outcomes are shown to readily outperform sequential design baselines. Significant differences in morphology and learned behavior are also observed when comparing co-designed single robot vs. co-designed multi-robot systems for similar operations.

\end{abstract}

\section{Introduction}
Inspired by natural systems, Multi-Robot systems (MRS) and Swarm Systems (SS) employ collective intelligence principles to exhibit emergent behavior to accomplish tasks that are beyond the capabilities of any single robot. 
Emergent behavior results from simple rules followed by each entity and their interaction with each other and their environment \cite{chen2022securing}. These interactions give rise to complex adaptive behaviors that are robust and efficient. Usually such collective behavior is not readily predictable (e.g., via scaling or simple equations) from individual behavior without the use of empirical evaluations via simulations. This is because appropriate design and behavior choices at the individual robot level can lead to collective performance that is greater than the sum of its parts 

Now, the physical design aka morphology of individual robots, including geometry and component choices w.r.t. sensors, actuators, computing, communication, etc., influence and constrain their operating envelope and functionalities. 
These design choices define the individual robot's capabilities (e.g., range, nominal power consumption, weight, sensing FoV, payload capacity, turning radius, etc.), and constrain the behavior space in which the robot can operate. 
On the other hand, the behavior (decision-system that perceives the environment and provides action) must align with the capabilities defined by its morphology. 
This creates a coupling of morphology and behavior individually. When working as a team, due to its task parallelization property, there are non-linear shifts in these constraints that affect its collective behavior. 
Realizing the true potential of swarm systems involves addressing formidable challenges regarding the design choices and behavior of the individual members. Even minor modifications in the design of individual robots might necessitate completely different behaviors. 

A common approach to designing swarm systems is by trial and error \cite{salman2019concurrent}.
The alternate method is the automated design approach, where the behavior is formulated as an optimization problem to be solved \cite{birattari2019automatic,mendiburu2022automode,ligot2022toward,kuckling2020automode,hasselmann2018automatic,francesca2014experiment}. 
These methods optimize the behavior of the individual robots using evolutionary methods and Reinforcement Learning (RL) methods to find the optimal behavior of individual robots that leads to the desired collective performance
\cite{trianni2008evolutionary}, \cite{lipson2005evolutionary}, \cite{christensen2006evolving}, \cite{hasselmann2021empirical}. 
By optimizing or prescribing the morphology first (as is typical), the capability space is inherently confined without considering the behavioral space, leading to a sub-optimal emergent behavior. 
The intricate interplay between \textit{morphology} and \textit{behavior}  must be carefully crafted together to explore how efficiently the swarm as a whole can achieve a desired collective behavior. 

There is a notable body of work on concurrent design or \textit{co-design} of morphology and behavior for individual robots ~\cite{sims1994evolving,weel2014robotic,khazanov2013exploiting,cheney2013unshackling,komosinski2009evolving,bongard2011morphological,gupta2021embodied,Schaff2018JointlyLT} and most of these methods use evolutionary approach, which however suffers from computational inefficiency and consider only the bounds of morphology space without taking geometric constraints into consideration.
There is limited literature on co-design in multi-robot systems \cite{watson2015deriving,salman2019concurrent}.
Most of these works are based on common simpler multi-robot problems such as foraging, aggregation, and formation. 
there is also a lack of computational frameworks for co-design that allow better understanding of how swarm systems compare with single-robot systems in terms of performance and how that relates to difference in morphology or behavior.  
To address these gaps, this paper proposes a computational framework that enables co-optimization of morphology and behavior of individual robots in a swarm or MRS to maximize collective performance, while also allowing compare/contrast analysis of single vs. swarm for a given problem. 
Here, we utilize our previously proposed concept of artificial-life-inspired talent metrics \cite{zeng2022efficient,krisshnakumar2024towards} that are physical quantities of interest, reflective of the capabilities of an individual robotic system. 
Talent metrics represent a compact yet physically interpretable parametric space that connects the behavior space and morphology space. We use this to decompose the morphology-behavior co-optimization into a sequence of talent-behavior optimization problems that can effectively reduce the overall search space (for each individual problem) with marginal compromise in the ability to find optimal solutions. In other words, the decomposition approach presented here is nearly lossless, i.e., a solution that can be found otherwise with a brute-force nested optimization approach to co-design will also exist in the overall search space spanned by our decomposed co-design approach (albeit assuming that each search process is ideal). 
We also propose a novel talent-infused policy gradient method to concurrently optimize the talents and learn the behavior.

To study operationally relevant behavior in this context, here we use a decentralized Multi-Robot Task Allocation (MRTA) problem, which finds applications in a wide range of real-world scenarios, some of which are search and rescue, disaster response, last-mile delivery, space exploration, and precision agriculture \cite{ghassemi2019decmrta,ju2022review,yliniemi2014multirobot}. 
In this paper, we consider a flood response scenario in which a group of UAVs collectively supply emergency packages throughout the environment. In our previous work, we proposed a graph capsule network-based RL policy for sequential task selection in such MRTA problems \cite{capam_mrta} which demonstrated superior performance compared to other baseline methods and proved to be scalable in terms of task space \cite{10.1115/1.4065883}. Therefore, it is adopted here to guide the behavior of the multi-robot system, which will now be co-optimized alongside the morphology of the individual robots, specifically UAVs in this scenario.

Thus, the primary contributions of this paper are as follows:
\textbf{1)} Present a new formulation and decomposed solution approach to concurrent (optimal) design of the morphology and learning-based behavior of multi-robot systems that are significantly more efficient than a nested co-design approach.
\textbf{2)} Develop an extension of the policy architecture used to embody the behavior (decisions) of robots in MRTA to also include (morphology-dependent) talents that can be simultaneously optimized through a policy gradient process.
\textbf{3)} Implement this new co-design approach to a flood response-inspired MRTA problem to identify and analyze the distinct morphology/behavior combinations obtained when using a single robot vs. using a multi-robot team (comprised of relatively simple individual robots) 
.
In section \ref{sec:codesign}, we present the co-design problem formulation, and section \ref{sec:MRTA} presents the learning-based MRTA planning approach that encompasses the behavior of the robots. Subsequently, the case study and its results are presented in Section \ref{sec:casestudy}, followed by concluding remarks in Section \ref{sec:conclusion}.
\section{Co-Design Framework}\label{sec:codesign}
\begin{figure*}
\centering
    \includegraphics[width=0.87\linewidth]{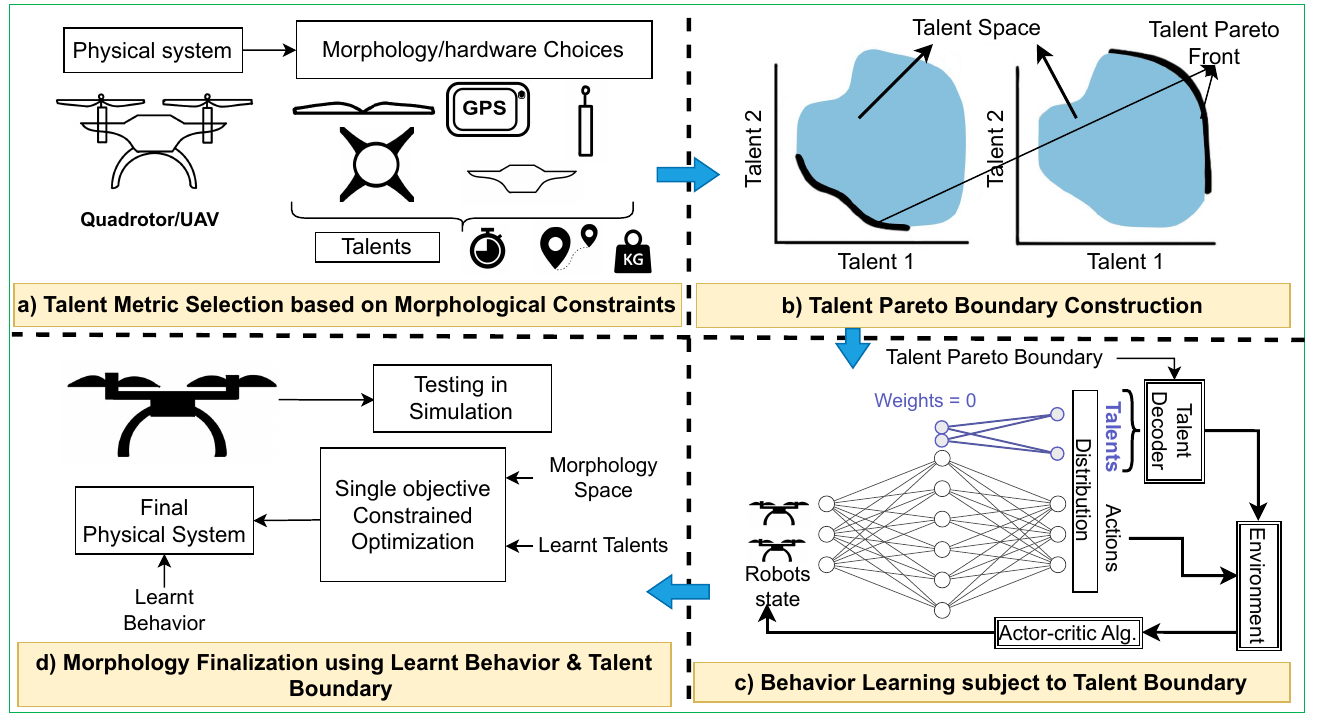}
    \caption{Flowchart of our co-design framework; a) Morphology and its dependent talent parameters are derived; b) Based on the talents, a Pareto front is created; c) The Talent-infused policy-gradient method is used to train the associated behavior and talents; d) Final morphology is obtained via constrained optimization subject to the learnt talent and behavior.}
  \label{fig:framework}
\end{figure*}
Consider a disaster response scenario in which a team of Unmanned Aerial Vehicles (UAVs) is deployed to deliver emergency relief supplies. 
The set of morphological variables, including physical form/geometry, component choices, and their physical properties (such as the motor and propeller sizes) can be expressed as $\mathbf{X}_\text{M}$. 
This vector comprises values $[X_1, X_2, \ldots X_n$], each corresponding to a distinct morphological variable. 
These robots follow a policy or behavior denoted by $\boldsymbol\Phi$ (representing policy parameters) to efficiently plan their mission. 
The collective performance based on this behavior can be represented by $f_\textbf{C}$, e.g., expressing metrics such as the number of packages delivered. 
The primary objective of co-design is to maximize the collective performance $f_\textbf{C}$ by simultaneously optimizing the morphological variables and the behavior while subject to geometric and other behavioral constraints. The optimization problem can thus be expressed as,
\begin{equation}
\footnotesize
\begin{aligned}
&\textrm{Max:} & & f_\textbf{C}(\mathbf{X}_\text{M}, \boldsymbol\Phi )\\
&\textrm{S. t.:} & &\mathbf{X}_{\texttt{min}} \leq \mathbf{X}_M \leq \mathbf{X}_{\texttt{max}}, \quad
\boldsymbol\Phi_{\texttt{min}} \leq \boldsymbol\Phi \leq \boldsymbol\Phi_{\texttt{max}}, \quad
 g(\mathbf{X}_\text{M}) \leq 0 
\end{aligned}
\label{eq:objfun_morph_1}
\end{equation}
where $g(\mathbf{X}_\text{M}) $ represent purely morphology-dependent constraints (e.g., geometric conflicts and component incompatibilities), and $[\mathbf{X}_{\texttt{min}},\mathbf{X}_{\texttt{max}}]$ and $[\boldsymbol\Phi_{\texttt{min}},\boldsymbol\Phi_{\texttt{max}}]$ are respectively the (lower, upper) bounds on the morphological and behavioral (policy) parameters. Figure \ref{fig:framework} depicts the four steps involved in our proposed co-design framework, which are explained in the later subsections. 

\subsection{Talent Metric Selection based on Morphological Constraints}\label{sec:talent_selection} 
During the mission, at each decision-making instance, i.e., after a package is delivered (one task completed), and the next location is determined, the robots have to consider the feasibility of proceeding to that location and completing the task. This includes assessing several factors such as the robot's remaining payload and its remaining range (ensuring it possesses enough battery to at least reach the location and return to depot afterward).
These factors, which influence collective behavior, are bounded by the capability of individual robots such as max flight range and max payload capacity, which are in turn dependent on morphology. Such capabilities are used here as ``\textit{talent}" metrics, as given by:

\begin{equation}\label{eq:talent_mapping}\footnotesize  \mathbf{Y}_{\texttt{TL}}=\left[Y_{\texttt{TL},1},Y_{\texttt{TL},2},\ldots,Y_{\texttt{TL},m}\right]
    =f_\text{M}(\mathbf{X}_\text{M})
\end{equation} 
where $f_\text{M}$ represents the function that maps candidate morphology variable vector to the set of $m$ talent metrics. Typically, in model-based design, such talent metrics or properties are computed using computational analysis models. 

Identifying talent metrics should follow four principles: 
1) The talent metrics must be solely a function of morphology (not affected by behavior). 
2) Talent metrics should exhibit the monotonic goodness property, meaning that for each metric, there should be a consistent direction of improvement (either the greater the better, or the smaller the better). 
3) Talent metrics should be collectively sufficient in computing state transitions of the system (for the given behavior context), and in determining the impact of morphology on behavior choices, meaning there cannot be a case where constraints or bounds on behavior can change with a fixed value of $\mathbf{Y}_{\texttt{TL}}$. 
4) Talent metrics should satisfy the basic multi-objective search property, i.e., they must conflict with each other in at least part of the (morphology) design space 

By adhering to these principles for identifying the talent metrics, we can reduce the computationally burdensome morphology-behavior co-optimization problem to a sequence of \textbf{1)} a multi-objective optimization to find the best trade-off talents, aka talent Pareto, and \textbf{2)} a talent/behavior co-optimization subject to not violating the determined talent Pareto. To elaborate, the second optimization must ensure that talent combinations that are beyond (or dominates) the talent Pareto front are not chosen during this process, since such combinations are in principle infeasible to achieve within the allowed morphological design space. Note that, in most robotic or complex engineering systems the dimensionality of $\mathbf{X}_\text{M}$ is usually much larger than that of $\mathbf{Y}_{\texttt{TL}}$ (the morphology space is considerably larger than the talent space), and thus this approach is also expected to enable searching a lower dimensional space during the co-optimization. 
This talent/behavior co-optimization process can be expressed as:

\begin{equation}
\footnotesize
\begin{aligned}
\textrm{Max:} & \quad f_{\mathbf{C}}(\mathbf{Y}_{TL}, \boldsymbol\Phi) \\
\textrm{S. t.:} & \quad {\text{min}}(Y_{\texttt{TL,1}}) \leq Y_{\texttt{TL,1}} \leq {\text{max}}(Y_{\texttt{TL,1}}) \\
& \quad Q(0.05 | Y_{\texttt{TL,1}}, \cdots, Y_{\texttt{TL,{i-1}}}) \leq Y_{\texttt{TL,i}} \leq Q(0.95| Y_{\texttt{TL,1}}, \cdots, Y_{\texttt{TL,{i-1}}}) \\
& \quad \forall ~i \in {2, \dots, m-1} \\
& \quad\boldsymbol\Phi_{\text{min}} \leq \boldsymbol\Phi \leq \boldsymbol\Phi_{\text{max}} \\
\end{aligned}
\label{eq:objfun_morph_2}
\end{equation}
Here, $Q$ represents the quantile regression model, and for every talent metric $Y_{\texttt{TL,i}}$ except the first one (i.e., $\forall i>1$), we progressively capture the 5th and 95th percentile values conditioned on  ($Y_\texttt{TL,1}, \cdots , Y_{\texttt{TL,i-1}}$) to use it as a lower bound and upper bound of $Y_{\texttt{TL,i}}$, respectively. For the first talent variable, we can directly acquire the bounds using the Pareto points. During co-optimization, allowed talent values must satisfy these bound constraints estimated thereof.  
 
\subsection{Talent Pareto Boundary Construction}\label{sec:talent_pareto_generation}

Consider a set of two talent metrics. Figure \ref{fig:framework}~b) represents the feasible talent space, and based on example min-min and max-max scenarios, the lower left and upper right boundaries of this space respectively represent the Pareto front. So, for say a max-max scenario (e.g., consisting of the flight range and nominal speed of the UAV), any point further North-East of the upper right boundary is not achievable, i.e., gives an infeasible morphology candidate. In other words, the Talent Pareto not only bounds all feasible combinations of flight range and speed, but also allows us to pick best trade-off (non-dominated) combinations and ignore dominated ones -- thus both constraining and reducing the search space of candidate talent combinations to consider downstream. 
Now, two steps are needed to identify and parametrically model this talent (Pareto) boundary:

\textbf{i)\textit{Multi-talent optimization}:}
A set of best trade-off talent combinations  (Pareto solutions) can be obtained by solving the following multi-objective optimization problem, e.g., using a standard genetic algorithm. 

\begin{equation}
\footnotesize
\begin{aligned}
&\textrm{Max:} & & (Y_{\texttt{TL,1}}, \ldots,Y_{\texttt{TL,m}}) = f_\text{M}(\mathbf{X}_\text{M})\\
&\textrm{S. t.:} & &\mathbf{X}_{\texttt{min}} \leq \mathbf{X}_M \leq \mathbf{X}_{\texttt{max}} , \quad
g(\mathbf{X}_\text{M}) \leq 0
\end{aligned}
\label{eq:objfun_morph_3}
\end{equation}

\textbf{ii)\textit{ Modeling the Pareto front}:}
A parametric representation of the Pareto front, $f_{\text{S}}$, namely the $m$-th talent expressed as a mathematical function of the remaining talent metrics, can be obtained by using a surrogate model such as a polynomial response surface to fit the computed talent Pareto solutions, i.e.: 

\begin{equation} 
\footnotesize
    Y_{\texttt{TL},m} = f_{\text{S}} \left(Y_{\texttt{TL},1},\ldots,Y_{\texttt{TL},m-1} \right)
    \label{eq:surrugate_model}
\end{equation}

\subsection{Behavior Learning with Talent optimization}\label{sec:talent-infused_actorcritic}  
To generate the behavior (policy) model, we use the actor-critic method, while other standard policy gradient techniques can be exploited here as well. 
The structure of the policy model will depend on the nature of the behavior being learnt. A generic example of neural net based policy, aka the actor network, is shown in Fig.~\ref{fig:framework}~c)(black). 

\textbf{\textit{Talent-infused Actor-critic}:} To co-optimize the behavior policy along with the talents, subject to the talent Pareto boundary obtained in the previous step (Fig.~\ref{fig:framework}~b)), we introduce a second small 2-layer fully connected neural net called the talent network, as shown in blue color in \ref{fig:framework}~c. 
The talent network architecture includes biases that are randomly initialized and pass through a linear activation layer. The resulting values are then forwarded to the output layer, which has $m-1$ neurons with sigmoid activation, where $m$ is the number of talents. So, the talent network does not essentially have any input layers or inputs, and is defined by the biases in the first and outputs layers and weights connecting these two layers, which are the parameters optimized during training. 
This network is concatenated to the behavior (policy) network. 
Let's consider this combined network to be the new actor network. The policy of this actor network is given by $ \pi((a|s, \hat{Y}_{\texttt{TL,1}}, \cdots, \hat{Y}_{\texttt{TL,m-1}}); \theta) $, where $a$ is the behavioral action, $\hat{Y}_{\texttt{TL,1}}, \cdots, \hat{Y}_{\texttt{TL,m-1}}$ are the talent values from the talent network and $\theta$ indicates the parameters of the combined actor network. 

\textbf{\textit{Training Phase for talent-behavior co-optimization}:}
In RL, a common strategy for exploration involves sampling actions from a distribution. 
Here, since the talents are continuous, a Gaussian distribution can be utilized.  During the first step of each episode, we do a forward pass in the actor network (consisting of both the talent network and behavior network), followed by sampling from the distribution. The augmented output of the actor network is given by 

\begin{equation}
\footnotesize
    A_\theta(s_{t}) = (a_{t}, \hat{Y}_{\text{TL,1}}, \hat{Y}_{\text{TL, 2}}, \ldots, \hat{Y}_{\text{TL, m-1}}), 
    \quad \text{for all } t \in \{1, \ldots, T\}
    \label{eq:actor_policy}
\end{equation}
where $A_\theta(s_{t})$ signifies the output of actor policy at time step $t$ with input state $s_{t}$, and $a_{t}$ represents the action for state $s_{t}$, $\hat{Y}_{\texttt{TL,1}}, \hat{Y}_{\texttt{TL, 2}}, .., \hat{Y}_{\texttt{TL, m-1}}$ represents the talent values from 1 to $m-1$. 
These $m-1$ values are subsequently processed by a talent decoder, which scales them based on the upper and lower bounds of their respective talent metric. For the first talent metric, we get:
\begin{equation}
\footnotesize 
    {Y}_{\texttt{TL},1} = \hat{Y}_{\texttt{TL},1}(\max(\hat{Y}_{\texttt{TL,1}}) - \min(\hat{Y}_{\texttt{TL,1}}))  + \min(\hat{Y}_{\texttt{TL,1}})
\end{equation}
For remaining, 2nd to $m-1$, talents, we use the following equation: 
 
\begin{equation}
\begin{aligned}
        Y_{\texttt{TL},i} = &\ \hat{Y}_{\texttt{TL},i} \left( Q(0.95|Y_{\texttt{TL},1}, \ldots, Y_{\texttt{TL},i-1}) \right. 
     \left. - Q(0.05|Y_{\texttt{TL},1}, \ldots, Y_{\texttt{TL},i-1}) \right) + \\
    &\ Q(0.05|Y_{\texttt{TL},1}, \ldots, Y_{\texttt{TL},i-1}),~~\ \forall i \in \{2, \ldots, m-1\}
\end{aligned} 
\end{equation}
To obtain the last talent in the set, namely, $Y_{\texttt{TL,m}}$, we use the surrogate model created with eqn. \ref{eq:surrugate_model} in the previous step of our co-design approach. 
After deciding on actions and talents, we input these into the simulation. Robots are created using these talents. Note that for the MDP computations, the robot capabilities expressed as the talent set is necessary and sufficient to model or embody the robot agent in the simulation (their morphology doesn't need to be explicitly determined). Once the talent based robot has been defined, the computed action $a_{t}$ is taken to get rewards and new states, that are returned to the actor network. 
Crucially, after the first step of the episode, talents are not sampled from the distribution, since they are not input dependent. Moreover, changing the talent and thus the robot design during an episode would not be physically meaningful. Thus, only behavioral actions are forward propagated throughout the episode, i.e., the states and actions update with each step, as shown in the eqn \ref{eq:actor_policy}. 

The critic network, which primarily gets the states as input, is modified to receive state-talent values. Now, instead of calculating the state value, the critic network calculates the state-talent values. The new critic policy can be represented as $V(s_t,\hat{Y}_{\texttt{TL}}; w)$. 
The Temporal Difference (TD) error is then computed based on $\delta = r + \gamma V(s_{(t+1)},\hat{Y}_{\texttt{TL}}; w) - V(s_t, \hat{Y}_{\texttt{TL}}; w) $

Since the talent values remain the same throughout the episode, it is necessary that we collect experiences containing batches of episodes and update the actor and critic networks over this batch. The TD error can be used to update the critic to optimally estimate the state-talent value. Consequently, the actor network is updated to increase the probability of providing us with the optimal Talents and behavior (actions based on states). Once the training converges, the deterministic actor provides the optimal talents ($\mathbf{Y}_{\texttt{TL}}^*$), and the behavior policy.





\subsection{Morphology Finalization}\label{sec:morphology_finalization}
Utilizing the optimized or learnt talent metrics $\mathbf{Y}_{\texttt{TL}}^*$, we determine the final robot morphology through another single objective optimization process, as shown in fig. \ref{fig:framework} d). The goal of this optimization is to now explicitly find the morphology that corresponds to (as closely as possible) the optimal talent metrics obtained in the previous step. This optimization can be expressed as, 

\begin{equation}
\footnotesize
\begin{aligned}
&\textrm{Min:} & & f_f=||\mathbf{Y}_{\texttt{TL}}(\mathbf{X}_M)-\mathbf{Y}_{\texttt{TL}}^*|| \\
&\textrm{S. t.:} & &\mathbf{X}_{\texttt{min}} \leq \mathbf{X}_M \leq \mathbf{X}_{\texttt{min}} , \quad g(\mathbf{X}_\text{M}) \leq 0
\end{aligned}
\label{eq:morphology_finalization}
\end{equation}
Any standard non-linear constrained optimization solver can be used here. In this paper, we use a Particle Swarm Optimization implementation \cite{chowdhury2013mixed}. 


\section{Multi-Robot Task Allocation for Flood Response}\label{sec:MRTA}
In this work, we focus on a multi-unmanned aerial vehicle (UAV) flood disaster response problem adopted from \cite{ghassemi2021multi,10.1115/1.4065883}, which we refer to as {MRTA-Flood}. It consists of $N_{T}$ task locations in a flood-affected area waiting for a survival package to be delivered by a team of  $N_{R}$ UAVs. Here, the goal is to drop survival packages to as many task locations as possible before the water level rises significantly, submerging all the locations. 
We assume that each location requires just one survival package. 
The predicted time at which a location $i$ gets completely submerged ($\tau_i$) is considered as the deadline of the task $i$, by which time that task must be completed.  
Each UAV has a max package (payload) capacity, max flight speed and max flight range, which comprise the set of talents. 
We consider a \textit{decentralized asynchronous} decision-making scheme.
The following assumptions are made: 1) All UAVs are identical and start/end at the same depot; 2) The location $(x_i,y_i)$ of task-$i$ and its time deadline $\tau_i$ are known to all UAVs; 3) Each UAV can share its state and its world view with other UAVs; and 4) A linear charging model with a charging time from empty to full range being 50 minutes, the charging happens every time the UAV visits the depot.

\subsection{MRTA-Flood  Problem Formulation}
\label{sec:ProblemFormulation}
Here, we present a summary of the Markov Decision Process (MDP) formulation of this multi-UAV flood-response problem.
\begin{table}
\scriptsize
\caption{The state and action parameters of the graph learning policy for MRTA-Flood}\label{tab:mdp}
\begin{tabular}{|l|l|} \hline
\multicolumn{1}{|l}{} & Parameter \\ \hline
\multirow{10}{*}{States} & Task graph ($\mathcal{G}$) \\
 & current mission time ($t$) \\
 & current location of the robot ($x^{t}_{r}, y^{t}_{r}$) \\
 & remaining battery of robot $r$ ($\phi^{t}_{r}$) \\
 & capacity of robot $r$ ($c^{t}_{r}$) \\
 & destination of its peers ($x_{k}, y_{k}, k \in [1, N_{R}], k \neq r$) \\
 & Remaining battery of peers ($\phi^{t}_{k}, k \in [1, N_{R}], k \neq r$) \\
 & Capacity of peers ($c^{t}_{k}, k \in [1, N_{R}], k \neq r$) \\
 & Destination time of peers($t^{next}_{k}, k \in [1, N_{R}], k \neq r$) \\
 & Talents ($\hat{Y}_{\texttt{TL,1}}$ and $\hat{Y}_{\texttt{TL,2}}$) \\ \hline
Actions & Task to allocate (${0,\ldots,N_{T}}$) \\ \hline
\end{tabular}
\end{table}
\textbf{MDP over a Graph:}

The MRTA-Flood problem involve a set of nodes/vertices ($V$) and a set of edges ($E$) that connect the vertices to each other, which can be represented as a complete graph $\mathcal{G} = (V, E, \Omega)$, where $\Omega$ is a weighted adjacency matrix. Each node represents a task, and each edge connects a pair of nodes. For MRTA with $N_{T}$ tasks, the number of vertices and the number of edges are $N_{T}$ and $N_{T}(N_{T}-1)/2$, respectively. Node $i$ is assigned a 3-dimensional feature vector denoting the task location and time deadline, i.e., $\rho_i=[x_i,y_i,\tau_i]$ where $i \in [1, N_{T}]$. Here, the weight of the edge between nodes $i$ and $j$ is $\omega_{ij}$ ($\in \Omega$), which can be computed as $\omega_{ij} = 1 / (1+\sqrt{(x_{i}-x_{j})^{2} + (y_{i}-y_{j})^{2} + (\tau_i - \tau_j)^2})$, where $i, j \in [1,N_{T}]$.
 
\begin{table}
    \centering
        \caption{UAV talent metrics and design variables obtained in co-design compared with baseline designs for MRTA and single robot task allocation (SRTA)}
    \includegraphics[trim={0 6.4cm 0 0},clip,width=0.95\linewidth]{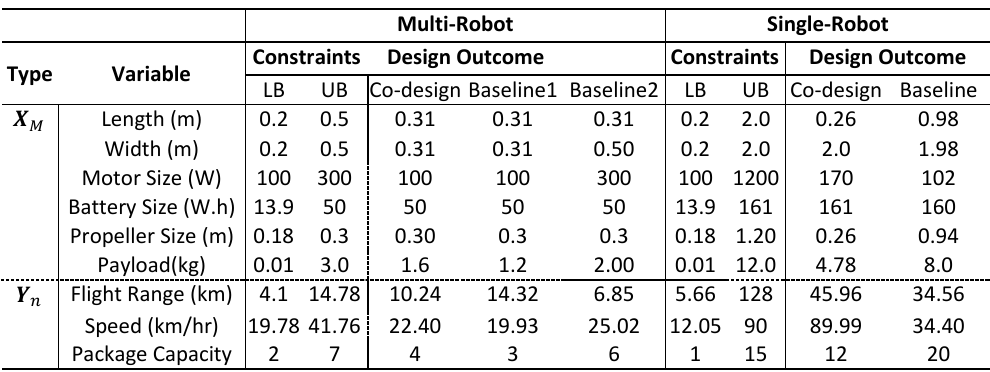}
    \label{tab:DVs}
\end{table}
The MDP defined in a decentralized manner for each individual UAV (to define its task selection process) can be expressed as a tuple $<\mathcal{S}, \mathcal{A}, \mathcal{P}_a, \mathcal{R}>$. The \textbf{State Space ($\mathcal{S}$)} consists of the task and peer robot properties and mission-related information. The \textbf{Action Space ($\mathcal{A}$)} is the index of the task selected to be completed next $\{0,\ldots,N_{T}\}$ with the index of the depot as $0$. The full state and action space is shown in table \ref{tab:mdp}. The (\textbf{Reward function ($\mathcal{R}$)}) is defined as
$10 \times N_{\text{success}}/N_{T}$, where $N_{\text{success}}$
is the number of successfully completed tasks and is calculated at the end of the episode. Since here we do not consider any uncertainty, the state transition probability is 1.

\subsection{Graph-Based Behavior Policy Network} 
\begin{figure*}[!ht] 
    \centering
    \includegraphics[width=0.7\linewidth]{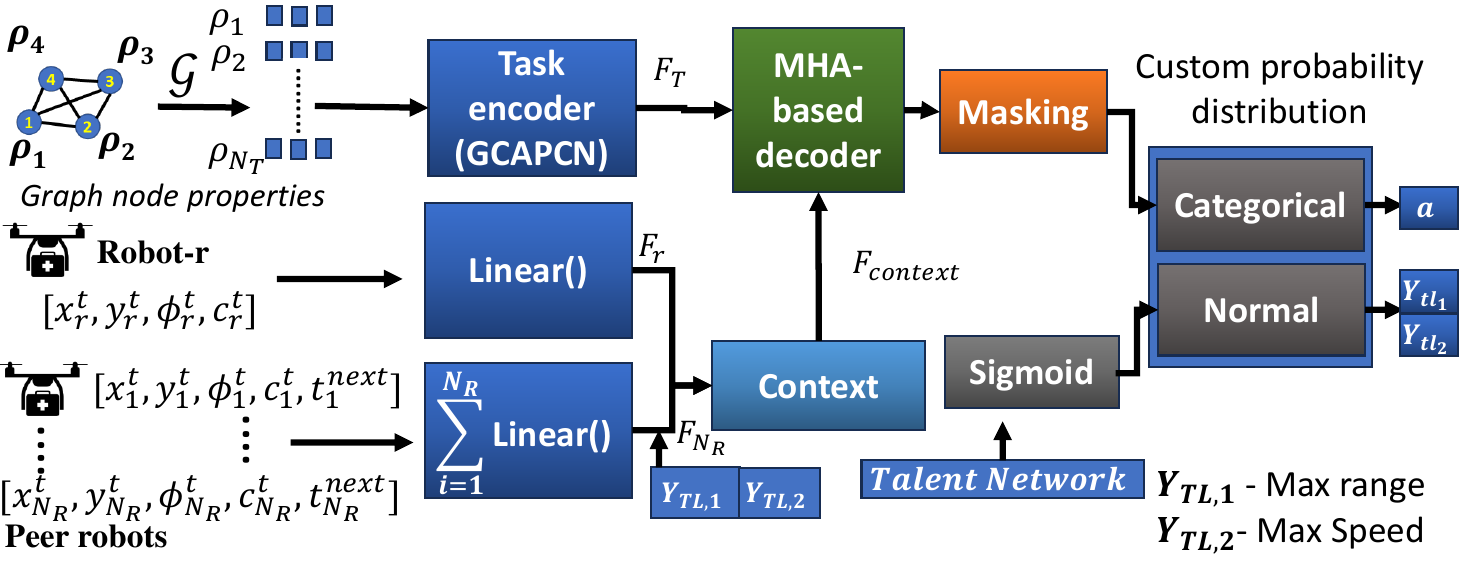}
    \caption{The overall policy network consists of the GCAPCN encoder, context encoding, the MHA-based decoder, and Talent network.}
    \label{fig:policy} 
\end{figure*}

Motivated by the generalizability and scalability benefits reported in \cite{capam_mrta,paul2023efficient,10.1115/1.4065883}, we construct a policy network based on specialized Graph Neural Networks (GNN) which maps the state information to an action. The behavior policy network consists of a Graph Capsule Convolutional Neural Network (GCAPCN) \cite{Verma2018} for encoding the graph-based state information (the task graph). The remaining state information, which includes the state of the robot tasking decision, the peer robots, and the maximum range ($\hat{Y}_{\texttt{TL,1}}$), and maximum speed ($\hat{Y}_{\texttt{TL,2}}$), are concatenated as a single vector and passed through a linear layer to obtain a vector called the context ($F_{\text{context}} \in \mathbb{R}^{h_{l} \times 1}$, where $h_{l}$ is the length of the context vector). The encoded and context information are then processed by a decoder to compute the actions, namely the probability of selecting each available task. Figure \ref{fig:policy} shows the overall policy network. 
Further details of the GCAPCN encoder and the MHA-based decoder can be found in our previous works \cite{capam_mrta}, \cite{paul2023efficient}, \cite{paul2022scalable}, and are thus not elaborated here.

\section{Case study - Results and Discussion}\label{sec:casestudy} 
This section showcases the implementation and results of each step of the co-design framework applied to the MRTA-Flood problem.

\subsection{Talent metrics and Pareto Front} 
For the MRTA-Flood problem, we consider a quadcopter with a Blended-Wing-Body (BWB) design integrated into an H-shaped frame \cite{zeng2022efficient}. 
The key morphological parameters that influence performance are the length and width of quadcopter arms, motor power, battery capacity, payload, and propeller diameters, although a much larger or more granular set of design variables can also be readily considered in future implementations. As stated earlier, the talent set comprises flight range ($Y_{\texttt{TL,1}}$), nominal speed ($Y_{\texttt{TL,2}}$), and the payload or package capacity ($Y_{\texttt{TL,3}}$) of the UAV. We consider each package to have 400 grams of emergency supplies. Computational underlying the design objective and constraint calculations (eq. \ref{eq:talent_mapping}) for this UAV can be found in \cite{zeng2022efficient}. 
\begin{figure*}
    \centering
    \includegraphics[width=.75\linewidth]{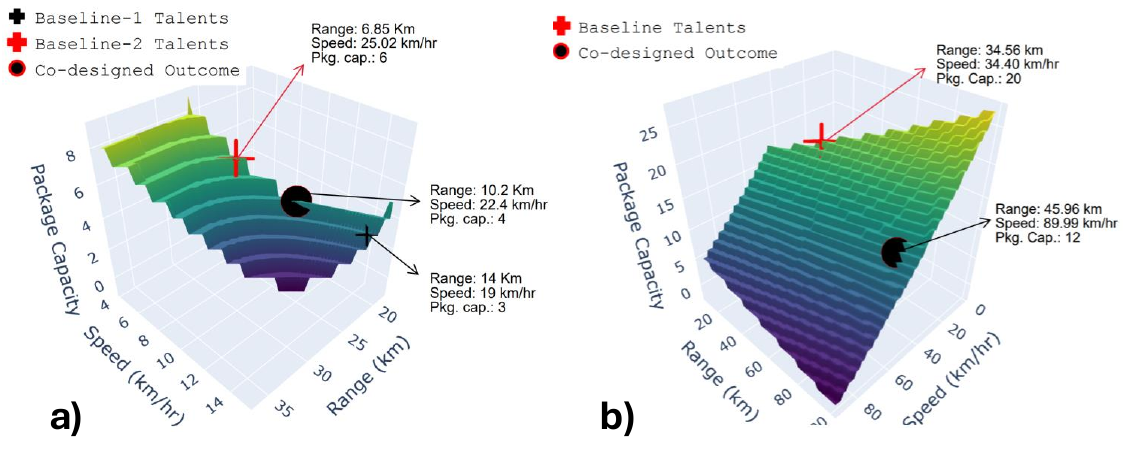}
    \caption{Talent Pareto front approximated by polynomial regression; limits of talents captured with quantile regression. a) Pareto front for MRTA morphology constraints, b) Pareto front for SRTA morphology constraints}
    \label{fig:pareto_model} 
\end{figure*}
In order to identify the Pareto points as explained in section \ref{sec:talent_pareto_generation}, we utilize the NSGA-II (Non-dominated Sorting Algorithm II) solver. For robustness, the optimization process involved conducting six separate runs, with each run consisting of a population of 120 and 40 generations each. 
Subsequently, the Pareto points obtained through six runs are subjected to another final non-dominated sorting process to acquire the final set of Pareto points.
After the final sorting process, we identified a total of $289$ Pareto solutions.
Finally, to capture and model the Pareto front, we utilize 2D Quadratic regression, considering Range and Speed as independent variables and package capacity as the dependent variable, $Y_{\texttt{TL},m}$. The resulting Pareto front is shown in figure \ref{fig:pareto_model}~a). \vspace{-.5cm}


\subsection{Behavior Learning subject to Talent Boundary}\label{subsec:behavior_learning}
We use the Stable-baselines3 \cite{stable-baselines3} library to implement our custom policy and distribution as elaborated in section \ref{sec:MRTA}. 
The policy generates outputs for flight range ($\hat{Y}_{\texttt{TL,1}}$) and speed ($\hat{Y}_{\texttt{TL,2}})$, both in a range of [0,1]. 
The flight range is scaled using the upper and lower bounds of the flight range of the Pareto front given in table \ref{tab:DVs}. 
The speed output $\hat{Y}_{\texttt{TL,2}}$ undergoes scaling based on the 5th and 95th percentile values (upper and lower limits) of speed conditioned on the flight range.
The number of packages is estimated using the polynomial regression created with flight range and speed as inputs. 
The behavioral action ($a$) is also determined as a part of the policy, indicating the next task to complete.
The training area is 5 sq. km, the number of robots is 5, the task size is 50, and the total mission time is 2 hours,  which is kept fixed for ease of computation. 
For each episode, the depot, task locations, and the time deadline for each task are randomly generated across the environment. 

\begin{figure}
    \centering
    \includegraphics[trim={+0.1cm +0.4cm 0 0},width=1.0\linewidth]{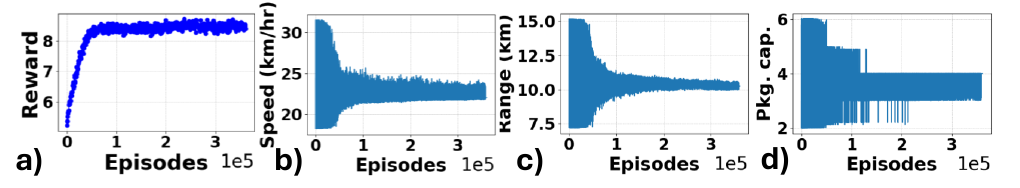}
    \caption{Training history for MRTA co-design policy (Talents and overall reward): (a) Reward, (b) Cruise speed, (c) Flight range, (d) Package Capacity}
\label{fig:overall_convergence_mrta}
\end{figure}
\vspace{-0.05cm}
The policy is trained using Talent-infused Proximal policy optimization (PPO) for approximately 350k episodes. Figure \ref{fig:overall_convergence_mrta} shows the convergence history of talents and rewards. During the initial part of the training, the standard deviation of the policy is higher, and as the training progresses and rewards (Figure \ref{fig:overall_convergence_mrta}~a) start to converge, the uncertainty of talents (Figure \ref{fig:overall_convergence_mrta}~b,c,d) reduces, signifying a stable learning process. 
The final cumulative standard deviation of the policy narrows down to 6.9\%, indicating a high level of precision and consistency in the learning outcomes. 
%
 
\subsection{Baseline Comparison}\label{subsec:baseline}
 
To compare our co-design policy's performance, we trained two baseline policies, and each has fixed talents: one possessing a higher package capacity and increased speed and the second baseline with a lower package capacity and higher range compared to our co-designed talents. 
The baseline talents are also selected from the Pareto front obtained through optimization (so they are competitive best trade-offs). The baselines represent typical automated sequential designs. 
Furthermore, by selecting the candidates from the Pareto front, we ensure that our co-design policy is benchmarked against design candidates that are better in one or more talents. 
 
\begin{figure}
    \centering
    \includegraphics[trim={+0.5cm +0.6cm 0 0}, width=0.95\linewidth]{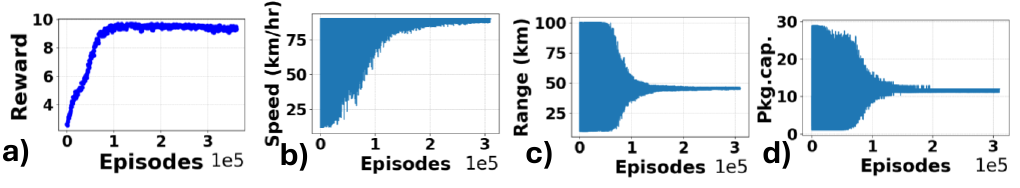}
    \caption{Training history for Single Robot Task Allocation (Talents and overall reward): (a) Reward, (b) Cruise speed, (c) Flight range, (d) Package Capacity}
\label{fig:overall_convergence_srta}
\end{figure}

\begin{figure}
    \centering
    \includegraphics[width=0.99\linewidth]{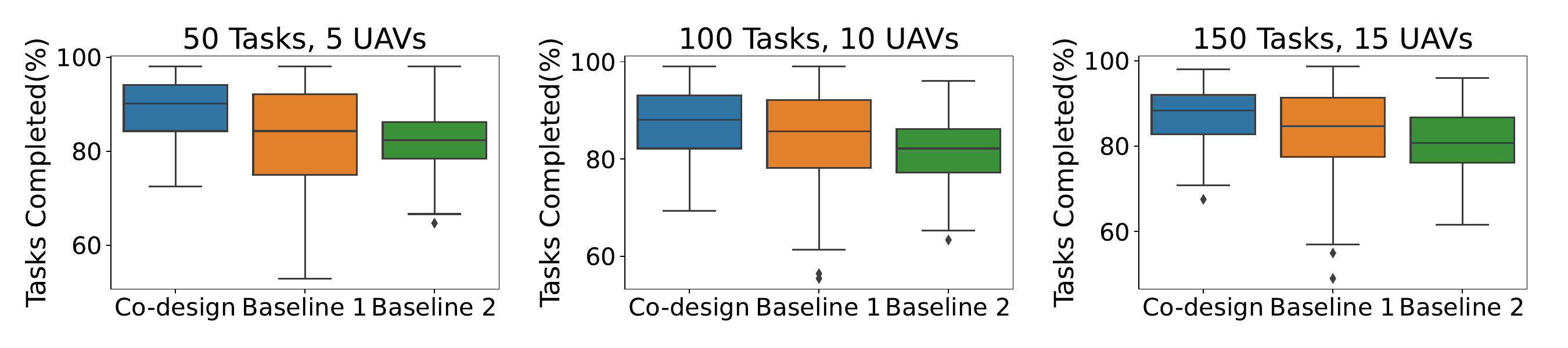}
    \caption{Multi-Robot Case: Task completion rate of co-designed policy and baseline policies with various task and UAV Scale}
    \label{fig:rewards_mrta} 
\end{figure}

Identical RL settings have been used throughout the experiments in this paper. 
The baseline behavior policies and the co-designed policy are evaluated with 3 different task sizes and robot counts across 250 episodes each. 
The task completion rate by each policy is compared in Figure \ref{fig:rewards_mrta}. 
In the training environment, which has 50 tasks and 5 UAVs, the co-designed policy demonstrated a median task completion rate of approximately 90\%, outperforming the baseline policies, which achieved around 83\% median task completion rate. As the environment was scaled to include 100 tasks with 10 UAVs and further to 150 tasks with 15 UAVs, the performance advantage of co-design over the baselines remained agnostic to scaling, further demonstrating the benefits of the co-design over sequential design. 

 
 \subsection{Single Robot Task Allocation}
  
A single robot task allocation (SRTA) co-design case study is also performed here, to provide insights with regards to: 1) At what scale of problems do co-designed multi-robot teams -- by virtue of task parallelism and emergent collective performance -- start providing benefits over a co-designed single robot deployment, where the latter is allowed a much larger or generous range of morphological choices (considering similar overall investment). 2) How the behavior/morphology combinations and inherent talents obtained from a single robot co-design differ from that of multi-robot co-design for the same operation. 

\subsection{Talent Boundaries and Behavior Learning} 
The upper bounds of our morphology variables are scaled 3-4 times in the single robot co-design baseline. Table \ref{tab:DVs} shows the upper and lower bounds of our morphology space for this single robot study. We obtained our Talent Pareto following the same method as before, and the resulting Pareto front is shown in Fig.~\ref{fig:pareto_model}~b). 
When compared with the Pareto front obtained for multi-robot morphology settings, the single robot Pareto front differs significantly, indicating that the influence of morphology on capability space is non-linear. 
The convergence history of talents and the rewards for Talent-infused learning in the single robot case are shown in Fig.~\ref{fig:overall_convergence_srta}. 
The co-design policy converges to a higher speed rather than a higher payload or flight range. A single UAV needs the speed to go to multiple locations and complete the tasks, while an appropriate balance between the number of packages it can carry and range is also necessary. 
A fixed design baseline policy is trained using talents from the single UAV Pareto front that have a higher payload than the single UAV co-designed talents. Both the baseline and optimized talents are shown in table \ref{tab:DVs}. 
In testing settings similar to the MRTA-Flood Problem, the single-robot co-designed policy surpasses the multi-robot policy in the training environment. However, its performance drops to 65\% when the number of tasks double and falls below 50\% when the number of tasks triple.
Since scalability is an essential component in any task allocation problem, when the number of tasks is changed, multi-robot systems provide a clear advantage. This hypothesized benefit remains evident even under co-designed outcomes (which arguably bring out the near-best of both worlds, single vs. multi-robot systems). 


\subsection{Final Morphology} 
The final morphologies for both the single-robot task allocation and multi-robot task allocation problems are provided in table \ref{tab:DVs}. While the upper bounds in morphology for a single robot system are scaled 3 to 4 fold for each variable, the optimized talents do not utilize the full bounds for most parameters. Interestingly, certain variables, such as the length and propeller size, were optimized to dimensions even smaller than the morphology observed in multi-robot configurations. In order to perform a more direct comparison of single-sophisticated and multi-(simple)-robot performance and how the behavior/morphology combinations offer distinct, not necessarily intuitive, trade-offs, an anchor is needed to equate the overall investment across these cases, e.g., total cost or mass (pertinent in space applications), and would be investigated in future work. 

\subsection{Computing Costs Analysis}
Our talent-behavior learning was performed in a workstation with Intel CPU-12900k (24 Threads), NVIDIA 3080ti, and 64 GB of RAM. The computation times for each step in our co-design framework for MRTA-Flood problem are: 6.7 minutes for 6 runs of NSGA-II to obtain talent Pareto solutions, just 3.5 seconds for generating the Pareto boundary regression model, 9 hours 57 minutes to train the talent-infused Actor-Critic policy, and 2.3 minutes for morphology finalization with MDPSO \cite{chowdhury2013mixed}. Overall, our co-design framework incurs a total computational cost of approximately 10 hours and 5 minutes. 
Using the policy training time (of 6 hours 49 minutes) with fixed morphology (namely the inner loop search) as reference, a nested co-design is estimated to take 272 hours if using NSGA-II for solving the outer level optimization. 

\begin{figure}
    \centering
    \includegraphics[width=0.99\linewidth]{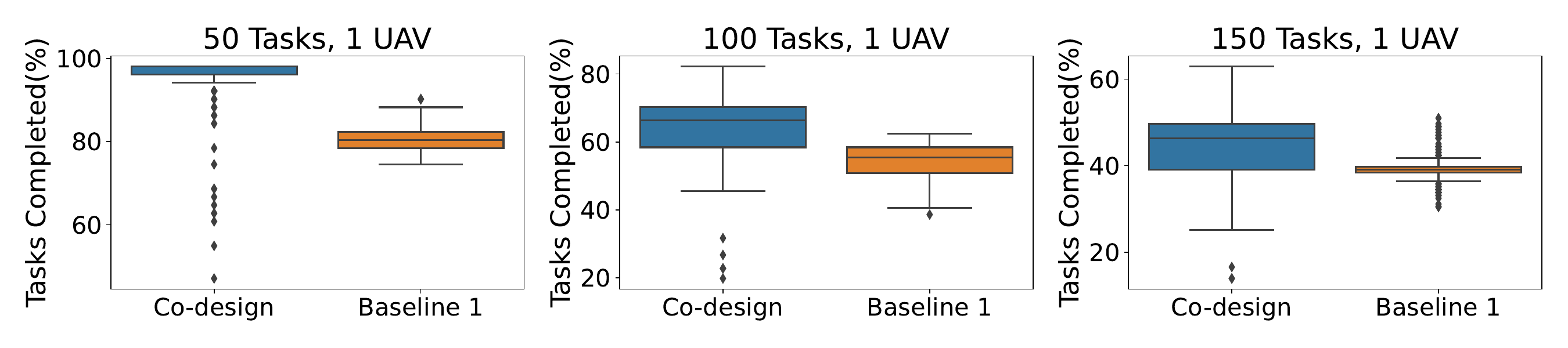} 
    \caption{Single Robot Case: Task completion rate of co-designed policy and baseline policy with various task counts}
    \label{fig:rewards_srta}
\end{figure}


\section{Conclusion}\label{sec:conclusion} 
In this paper, we introduced a new computational framework to concurrently design the learning-based behavior and morphology of individual robots in a multi-robot system, applied to the multi-robot task allocation context. Regression-based representation of the relation between the best trade-off talent choices (that represent robot capabilities), and formulation of a talent-infused actor critic policy, play key roles in enabling this new framework, with significant gains in computing efficient compared to a vanilla nested co-design approach. 
Applied to a multi-UAV flood response scenario, with the individual UAV behavior expressed by a graph neural network, the co-designed UAV team readily outperforms two sequential design baselines in terms task completion performance evaluated over unseen test scenarios. The framework also provides transparent insights into when a multi-UAV team becomes more beneficial compared to using a stand-alone more capable single UAV, and what morphological trades-offs occur between these two options. In its current form, the talent metrics must be purely functions of morphology, as well as be collectively sufficient to simulate the state transition underlying the robot behavior, which might be challenging to apply in settings with more complex robot/environment interactions. Future work can thus investigate talent representations that alleviate these assumptions, and thus allow wider application of the proposed co-design concept.

\bibliographystyle{IEEEtran}
\bibliography{refs}

\end{document}